# Assessing Sentiment Strength in Words Prior Polarities

*Lorenzo Gatti*[1]   *Marco Guerini*[1]
(1) TRENTO-RISE, 38123 Povo, Trento, Italy
lorenzo.gatti@trentorise.eu, marco.guerini@trentorise.eu

## Abstract

Many approaches to sentiment analysis rely on lexica where words are tagged with their prior polarity - i.e. if a word out of context evokes something positive or something negative. In particular, broad-coverage resources like SentiWordNet provide polarities for (almost) every word. Since words can have multiple senses, we address the problem of how to compute the prior polarity of a word starting from the polarity of each sense and returning its *polarity strength* as an index between -1 and 1. We compare 14 such formulae that appear in the literature, and assess which one best approximates the human judgement of prior polarities, with both regression and classification models. *To appear at Coling 2012*

# 1 Introduction

Many approaches to sentiment analysis use bag of words resources - i.e. a lexicon of positive and negative words. In these lexica, words are tagged with their prior polarity, that represents how a word is perceived out of context, i.e. if it evokes something positive or something negative. For example, *wonderful* has a positive connotation - prior polarity -, and *horrible* has a negative prior polarity. The advantage of these approaches is that they don't need deep semantic analysis or word sense disambiguation to assign an affective score to a word and are domain independent (so, less precise but portable).

Unfortunately, many of these resources are manually built and have a limited coverage. To overcome this limitation and to provide prior polarities for (almost) every word, other broad-coverage resources - built in a semi-automatic way - have been developed, such as SentiWordNet (Esuli and Sebastiani, 2006). Since words can have multiple senses and SentiWordNet provides polarities for each sense, there is the need for "reconstructing" prior polarities starting from the various word senses polarities (also called 'posterior polarities'). For example, the adjective *cold* has a posterior polarity for the meaning "having a low temperature" - like in "*cold* beer" - that is different from the polarity in "*cold* person" that refers to "being emotionless". Different formulae have been used in the previous literature to compute prior polarities (e.g. considering the posterior polarity of the most frequent sense, averaging over the various posterior polarities, etc.), but no comparison or analysis has ever been tried among them. Furthermore, since such formulae are often used as baseline methods for sentiment classification, there is the need to define a state-of-the-art performance level for approaches relying on SentiWordNet.

The paper is structured as follows: in Section 2 we briefly describe our approach and how it differentiates from similar sentiment analysis tasks. Then, in Section 3 we present SentiWordNet and overview various formulae appeared in the literature, which rely on this resource to compute words prior polarity. In Section 4 we introduce the ANEW resource that will be used as a gold standard. From section 5 to 7 we present a series of experiments to asses how good SentiWordNet is for computing prior polarities and which formula, if any, best approximates human judgement. Finally in Section 8 we try to understand whether the findings about formulae performances can be extended from the regression framework to a classification task.

# 2 Proposed Approach

In this paper we face the problem of assigning affective scores (between -1 and 1) to words. This problem is harder than traditional binary classification tasks (assessing whether a word - or a fragment of text - is either *positive* or *negative*), see (Pang and Lee, 2008) or (Liu and Zhang, 2012) for an overview. We want to asses not only that *pretty*, *beautiful* and *gorgeous* are positive words, but also that *gorgeous* is more positive than *beautiful* which, in turn, is more positive than *pretty*. This is fundamental for tasks such as affective modification of existing texts, where not only words polarity, but also their strength, is necessary for creating multiple "graded" variations of the original text (Guerini et al., 2008). Some of the few works that address the problem of sentiment strength are presented in (Wilson et al., 2004; Paltoglou et al., 2010), however, their approach is modeled as a multi-class classification problem (*neutral*, *low*, *medium* or *high* sentiment) at the sentence level, rather than a regression problem at the word level. Other works, see for example (Neviarouskaya et al., 2011), use a fine grained classification approach too, but they consider emotion categories (*anger*, *joy*, *fear*, etc.), rather than sentiment strength categories.

On the other hand, even if approaches that go beyond pure prior polarities - e.g. using word bigram features (Wang and Manning, 2012) - are better for sentiment analysis tasks, there are tasks that are intrinsically based on the notion of words prior polarity. Consider for example the task of naming, where evocative names are a key element to a successful business (Ozbal and Strapparava, 2012; Ozbal et al., 2012). In such cases no context is given for the name and the brand name alone, with its perceived prior polarity, is responsible for stating the area of competition and evoking semantic associations.

## 3 SentiWordNet

One of the most widely used resources for sentiment analysis is SentiWordNet (Esuli and Sebastiani, 2006). SentiWordNet is a lexical resource in which each word is associated with three numerical scores: `Obj(s)`, `Pos(s)` and `Neg(s)`. These scores represent the objective, positive and negative valence of the entry respectively. Each entry takes the form `lemma#pos#sense-number`, where the first sense corresponds to the most frequent.

Obviously, different word senses can have different polarities. In Table 1, the first 5 senses of `cold#a` present all possible combinations: a negative score only (`cold#a#1` and `cold#a#2`), a positive and objective score only (`cold#a#5`, `cold#a#3`), and mixed scores (`cold#a#4`). Intuitively, mixed scores for the same sense are acceptable, like in "*cold* beer" vs. "*cold* pizza".

| PoS | Offset  | PosScore | NegScore | SynsetTerms |
|-----|---------|----------|----------|-------------|
| a   | 1207406 | 0.0      | 0.75     | cold#a#1    |
| a   | 1212558 | 0.0      | 0.75     | cold#a#2    |
| a   | 1024433 | 0.0      | 0.0      | cold#a#3    |
| a   | 2443231 | 0.125    | 0.375    | cold#a#4    |
| a   | 1695706 | 0.625    | 0.0      | cold#a#5    |

Table 1: First five *SentiWordNet* entries for `cold#a`

### 3.1 Prior Polarities Formulae

In this section we review the main strategies for computing prior polarities from the previous literature. All the prior polarities formulae provided below come in two different versions (except *uni* and *rnd*). Given a lemma with $n$ senses (`lemma#pos#n`), every formula $f$ is applied - separately - to all the $n$ posScores and negScores of the `lemma#pos`; once the prior polarities for positive and negative scores are computed according to that formula, to map the result to a single polarity score (that can be either positive or negative), the possibility is:

1. $f_m = MAX(|posScore|, |negScore|)$ - take the max of the two scores
2. $f_d = |posScore| - |negScore|$ - take the difference of the two scores

Both versions range from -1 to 1. So, considering the first 5 senses of `cold#a` in Table 1, the various formulae will compute `posScore(cold#a)` starting from the values $<0.0, 0.0, 0.0, 0.125, 0.625>$ and `negScore(cold#a)` starting from $<0.750, 0.750, 0.0, 0.375, 0.0>$. Then either $f_m$ or $f_d$ will be applied to `posScore(cold#a)` and `negScore(cold#a)` to compute the final polarity strength. For the sake of simplicity, we will describe how to compute the *posScore* of a given lemma, since *negScore* can be easily derived. In details *posScore* stands for `posScore(lemma#pos)`, while $posScore_i$ indicates the positive score for the $i^{th}$ sense of the `lemma#pos`.

**rnd.** This formula represents the baseline random approach. It simply returns a random number between -1 and 1 for any given `lemma#pos`.

**swrnd.** This formula represents an advanced random approach that incorporates some "knowledge" from SentiWordNet. It returns the `posScore` and `negScores` of a random sense of the `lemma#pos` under scrutiny. We believe this is a fairer baseline than *rnd* since SentiWordNet information can possibly constrain the values. A similar approach has been used in (Qu et al., 2008), even though the authors used the polarity information from the first match of the term in the SentiWordNet synsets list - i.e. ignoring senses order - rather than a pure random sense.

$$posScore = posScore_i \quad \text{where} \quad i = RANDOM(1, n) \tag{1}$$

**fs.** In this formula only the first (and thus most frequent) sense is considered for the given `lemma#pos`. This is equivalent to asking for `lemma#pos#1` SentiWordNet scores. Based on (Neviarouskaya et al., 2009), (Agrawal et al., 2009) and (Guerini et al., 2008) (that uses the $fs_m$ approach), this is the most basic form of prior polarities.

$$posScore = posScore_1 \tag{2}$$

**mean.** It calculates the mean of the positive and negative scores for all the senses of the given `lemma#pos`, and then returns either the biggest or the difference of the two scores. Used for example in (Thet et al., 2009), (Denecke, 2009) and (Devitt and Ahmad, 2007). An approach explicitly based on $mean_d$ is instead presented in (Sing et al., 2012).

$$posScore = \frac{\sum_{i=1}^{n} posScore_i}{n} \tag{3}$$

**senti.** This formula is an advanced version of the simple mean, and concludes that only senses with a score $\neq 0$ should be considered in the mean:

$$posScore = \frac{\sum_{i=1}^{n} posScore_i}{numPos} \tag{4}$$

where *numPos* and *numNeg* are the number of senses that have, respectively, a $posScore > 0$ or $negScore < 0$ value. It is based on (Fahrni and Klenner, 2008) and (Neviarouskaya et al., 2009).

**uni.** This method, based on (Neviarouskaya et al., 2009) extends the previous formula, by choosing the MAX between *posScore* and *negScore*. In case *posScore* is equal to *negScore* (modulus) the one with the highest weight is selected, where weights are defined as

$$posWeight = \frac{numPos}{n} \tag{5}$$

As mentioned before, this is the only method, together with *rnd*, for which we cannot take the difference of the two means, as it decides which mean (*posScore* or *negScore*) to return according to the weight.

**w1.** This formula weighs each sense with a geometric series of ratio 1/2. The rationale behind this choice is based on the assumption that more frequent senses should bear more "affective weight" than very rare senses, when computing the prior polarity of a word. The system presented in (Chaumartin, 2007) uses a similar approach of weighted mean.

$$posScore = \frac{\sum_{i=1}^{n}(\frac{1}{2^{i-1}} \times posScore_i)}{n} \quad (6)$$

**w2.** Similar to the previous one, this formula weighs each lemma with a harmonic series, see for example (Denecke, 2008):

$$posScore = \frac{\sum_{i=1}^{n}(\frac{1}{i} \times posScore_i)}{n} \quad (7)$$

## 4 ANEW

To asses how well prior polarity formulae perform, a gold standard is needed, with word polarities provided by human annotators. Resources, such as sentiment-bearing words from the General Inquirer lexicon (Stone et al., 1966) are not suitable for our purpose since they provide only a binomial classification of words (either *positive* or *negative*). The resource presented in (Wilson et al., 2005) uses a similar binomial annotation for single words; another potentially useful resource is WordNetAffect (Strapparava and Valitutti, 2004) but it labels terms with affective dimensions (*anger*, *joy*, *fear*, etc.) rather than assigning a sentiment score.

We then choose ANEW (Bradley and Lang, 1999), a resource developed to provide a set of normative emotional ratings for a large number of words (roughly 1 thousand) in the English language. It contains a set of words that have been rated in terms of pleasure (affective valence), arousal, and dominance. In particular for our task we considered the valence dimension. Since words were presented to subjects in isolation (i.e. no context was provided) this resource represents a human validation of prior polarities strength for the given words, and can be used as a gold standard. For each word ANEW provides two main metrics: $anew_\mu$, which correspond to the average of annotators votes, and $anew_\sigma$ that gives the variance in annotators scores for the given word. In the same way these metrics are also provided for the male/female annotator groups.

## 5 Dataset pre-processing

In order to use the ANEW dataset to measure prior polarities formulae performance, we had to align words to the `lemma#pos` format that SentiWordNet uses. First we removed from ANEW those words that did not align with SentiWordNet. The adopted procedure was as follows: for each word, check if it is present among SentiWordNet lemmas; if this is not the case, lemmatize the word with TextPro (Pianta et al., 2008) and check again if the lemma is present[1]. If it is not found, remove the word from the list (this was the case for about 30 words of the 1034 present in ANEW).

The remaining 1004 lemmas were then associated with the PoS present in SentiWordNet to get the final `lemma#pos`. Note that a lemma can have more than one PoS, for example,

---
[1] We didn't lemmatize words in advance to avoid duplications (for example, if we lemmatize the ANEW entry 'addicted', we obtain 'addict', which is already present in ANEW).

'writer' is present only as a noun (`writer#n`), while 'yellow' is present as a verb, a noun and an adjective (`yellow#v`, `yellow#n`, `yellow#a`). This gave us a list of 1494 words in the `lemma#pos` format. For each word, we tested the metrics described in Section 3.1 and annotated the results.

## 6 Evaluation Metrics

Given a formula for the prior polarities ($f$), we consider two different metrics to asses how well a formula performs on the ANEW dataset. The first metric is the Mean Absolute Error (*MAE*), that averages the error of the given formula on each ANEW entry. So given $n$ words $w$, we compute *MAE* as follows:

$$MAE = \frac{\sum_{i=1}^{n} |f(w_i) - anew_\mu(w_i)|}{n} \quad (8)$$

In multi-class classification problems a similar approach, based on Mean Squared Error (MSE), is used (based on a fixed threshold): if the strength of a sentence is high, classifying it as neutral (off by 3) is a much worse error than classifying it as medium (off by 1), (Wilson et al., 2004). The second metric, instead, tries to asses the percentage of successes of a given formula in assigning correct values to a word:

$$success = \frac{\sum_{i=1}^{n} [|f(w_i) - anew_\mu(w_i)| < \frac{1}{2} anew_\sigma(w_i)]}{n} = \frac{\sum_{i=1}^{n} [-\frac{1}{2} < zscore(w_i) < \frac{1}{2}]}{n} \quad (9)$$

*Success*, for a given word, is obtained when its *z-score* is between -0.5 and 0.5, i.e. the value returned by the formula, for the given word $w_i$, falls within one standard deviation $anew_\sigma(w_i)$ centered on the ANEW value. Assessing success according to the ANEW variance has the advantage of taking into account whether the given word has a high degree of agreement among annotators or not: for words with low variance (high annotator agreement) we need formulae values to be more precise. This approach is in line with other approaches on affective annotation that either assume one standard deviation (Grimm and Kroschel, 2005) or two (Mohammad and Turney, 2011) as an acceptability threshold and we chose the strictest one.

Finally, to capture the idea that the best approach to prior polarities is the one that maximizes success and minimizes error at the same time, we created a simple metric:

$$s/e = \frac{success}{MAE} \quad (10)$$

We decided to model the problem using *MAE* and *success* - rather than simply MAE (or MSQ) - in a regression framework, because we believe that apart from classification and ranking procedures (see (Pang and Lee, 2008) for an overview) traditional regression frameworks also cannot properly handle annotator's variability over polarity strength judgement (i.e. there is not a "true" sentiment value for the given word, rather an acceptability interval defined by the variability in annotators perception of prior polarity).

## 7 Analysis and Discussion

In Table 2, we present the results of the prior formulae applied to the whole dataset (as described in Section 5). In the following tables we report *success* and *MAE* for every formula; all formulae are ordered according to the $s/e$ metric. For the sake of readability, statistically significant differences in the data are reported in the discussion section. For *MAE* the significance is computed using Student's t-test. For *success* we computed significance using $\chi^2$ test.

| Metrics | $w2_m$ | $w1_m$ | $mean_m$ | $senti_m$ | $fs_m$ | $senti_d$ | $uni$ | $fs_d$ | $w2_d$ | $mean_d$ | $w1_d$ | $swrnd_d$ | $swrnd_m$ | $rnd$ |
|---|---|---|---|---|---|---|---|---|---|---|---|---|---|---|
| MAE | 0.377 | 0.379 | 0.378 | 0.379 | 0.390 | 0.381 | 0.380 | 0.390 | 0.380 | 0.382 | 0.382 | 0.397 | 0.400 | 0.624 |
| success | 32.5% | 32.5% | 32.3% | 32.3% | 33.1% | 31.7% | 31.5% | 32.1% | 31.2% | 30.9% | 30.9% | 30.5% | 30.6% | 19.9% |
| s/e | 0.864 | 0.858 | 0.856 | 0.852 | 0.848 | 0.834 | 0.830 | 0.825 | 0.820 | 0.810 | 0.810 | 0.767 | 0.765 | 0.319 |

Table 2: Function performances for all `lemma#pos`

| Metrics | $w2_m$ | $w1_m$ | $mean_m$ | $senti_m$ | $fs_m$ | $senti_d$ | $uni$ | $fs_d$ | $w2_d$ | $mean_d$ | $w1_d$ | $swrnd_d$ | $swrnd_m$ | $rnd$ |
|---|---|---|---|---|---|---|---|---|---|---|---|---|---|---|
| MAE | 0.381 | 0.384 | 0.383 | 0.385 | 0.405 | 0.388 | 0.386 | 0.404 | 0.387 | 0.390 | 0.390 | 0.418 | 0.422 | 0.638 |
| success | 33.1% | 32.9% | 32.7% | 32.6% | 34.0% | 31.6% | 31.2% | 32.3% | 30.6% | 30.2% | 30.2% | 29.3% | 29.6% | 21.1% |
| s/e | 0.868 | 0.857 | 0.854 | 0.846 | 0.840 | 0.815 | 0.809 | 0.800 | 0.791 | 0.774 | 0.774 | 0.702 | 0.700 | 0.331 |

Table 3: Function performances for `lemma#pos` with at least 1 SWN score $\neq 0$

We also focused on a particular subset to reduce noise, by ruling out "non-affective" words, i.e. those `lemma#pos` that have `posScore` and `negScore` equal to 0 for all senses in SentiWordNet - and for which the various formulae $f(w)$ always returns 0. Ruling out such words reduced the dataset to 55% of the original size to a total of 830 words. Results are shown in Table 3.

**SentiWordNet improves over Random**: the first thing we note - in Tables 2 and 3 - is that $rnd$, as expected, is the worst performing metric, while all other metrics have statistically significant improvements in results for both *MAE* and *success* (p<0.001). So, using SentiWordNet information for computing prior polarities increases the performance above baseline, regardless of the prior formula used.

**Picking up only one sense is not a good choice**: Interestingly $swrnd$ and $fs$ have very similar results which do not differ significantly (considering *MAE*). This means, surprisingly, that taking the first sense of a `lemma#pos` has no improvement over taking a random sense. This is also surprising since in many NLP tasks, such as word sense disambiguation, algorithms based on most frequent sense represent a very strong baseline[2]. In addition, *picking up one sense is also one of the worst performing strategies for prior polarities* and considering the mean error (*MAE*) the improvement over $swrnd_{m/d}$ and $fs_{m/d}$ is statistically significant for all other formulae (from p<0.05 to p<0.01).

**Is it better to use $f_m$ or $f_d$?**: The tables suggest that there is a better performance of prior formulae using $f_m$ over strategies using $f_d$ (according to $s/e$ such formulae rank higher). Still, on average, the *MAE* is almost the same (0.380 for $f_m$ formulae vs. 0.383, see Table 3). According to *success*, using the maximum of the two scores rather than the difference yields slightly better results (32.5% vs. 31.4%).

**Best performing formula, weighted average**: Best performing formulae on the whole dataset (according to $s/e$) are $w2_m$ and $w1_m$ (both on all words, in Table 2, and affective words in

---

[2]In SemEval 2010 competition, only 5 participants out of 29 performed better than the most frequent threshold (Agirre et al., 2010).

Table 3). In details, focusing on *MAE* and *success* metrics, and comparing results against $swrnd_d$ (the worst performing approach using SentiWordNet) we observe that: (i) considering *MAE*, significance level in Table 2 indicates that $w2_m$, $mean_m$, $w1_m$, $senti_m$ perform better than $swrnd_d$ (p<0.01). For Table 3 the same holds true but also including *uni* (p<0.01). (ii) Considering *success* the significance levels are milder, with p<0.05 and only for the best performing function on this metric ($fs_m$).

## 8 Prior Polarities and Classification tasks

Given the findings of the previous sections we can conclude that not all approaches to prior polarities using SentiWordNet are equivalent, and we manage to define a state-of-the-art approach. Still, since we conducted our experiments in a regression framework, we have to check if such findings also hold true for sentiment classification tasks, which are the most widely used. In fact, it is not guaranteed that significant differences in *MAE* or *success* are relevant when it comes to assessing the polarity of a word. Two formulae can have very different error and success rates on polarity strength assessment, but if they both succeed in assigning the correct polarity to a word, from a classification perspective the two formulae are equivalent.

In Table 4 we present the results of prior polarities formulae performance over a two-class classification task (i.e. assessing whether a word in ANEW is *positive* or *negative*, regardless of the sentiment strength). We also considered a classifier committee ($cc$) with majority vote on the other formulae (random approaches not included). Significance is computed using an approximate randomization test (Yeh, 2000) and formulae are ordered according to F1 metric. Note that in this task the difference between $f_m$ and $f_d$ is not relevant since both versions always return the same classification answer.

|  | $w2$ | $mean$ | $w1$ | $cc$ | $sentiful$ | $uni$ | $fs$ | $swrnd$ | $rnd$ |
|---|---|---|---|---|---|---|---|---|---|
| Precision | 0.712 | 0.708 | 0.706 | 0.705 | 0.703 | 0.698 | 0.687 | 0.666 | 0.493 |
| Recall | 0.710 | 0.707 | 0.705 | 0.704 | 0.702 | 0.699 | 0.675 | 0.653 | 0.493 |
| F1 | 0.711 | 0.707 | 0.706 | 0.705 | 0.702 | 0.698 | 0.681 | 0.659 | 0.493 |

Table 4: Precision, Recall and F1 in the classification task on positive and negative words.

Results are very similar to the regression case: all classifiers have a significant improvement over a random approach ($rnd$, p<0.001), and most of the formulae also over $swrnd$ with p<0.05. As before, $fs$ has no improvement over the latter (i.e. also in this case choosing the most frequent sense has the same poor performances of picking up a random sense). Furthermore $w2$, $mean$ and $w1$ - the best performing formulae in the regression case - have a stronger significance over $swrnd$ with p<0.01. This means that also for the classification task we can define a state-of-the-art approach for prior polarities with SentiWordNet based on (weighted) averages.

## 9 Conclusions

In this paper we have presented a series of experiments in a regression framework that compare different approaches in computing prior polarities of a word starting from its posterior polarities. We have shown that a weighted average over word senses is the strategy that best approximates human judgment. We have further shown that similar results holds true for sentiment classification tasks, indicating that also in this case that a weighted average is the best strategy to be followed.